\documentclass[runningheads]{llncs}

\usepackage{eccv}

\usepackage{tabularx}
\usepackage[normalem]{ulem}
\useunder{\uline}{\ul}{}
\newcolumntype{C}{>{\centering\arraybackslash}X}

\usepackage{eccvabbrv}

\usepackage{graphicx}
\usepackage{booktabs}
\usepackage{adjustbox}
\usepackage[misc]{ifsym}

\usepackage[accsupp]{axessibility}  

\usepackage{hyperref}

\usepackage{orcidlink}

\begin{document}

\title{Segmentation-guided Layer-wise Image Vectorization with Gradient Fills} 



\renewcommand{\thefootnote}{\ifcase\value{footnote}\or\Letter\or*\or\dagger\fi}

\author{Hengyu Zhou\orcidlink{0009-0001-1487-5313} \and
Hui Zhang{\textsuperscript{\,\Letter}}\orcidlink{0000-0001-6563-9890} \and
Bin Wang{\textsuperscript{\,\Letter}}\orcidlink{0000-0002-5176-9202}}

\footnotetext[1]{Corresponding authors.}

\authorrunning{H.~Zhou et al.}

\institute{School of Software, Tsinghua University, P. R. China
\email{zhouhy22@mails.tsinghua.edu.cn,\,\{huizhang,\,wangbins\}@tsinghua.edu.cn}}

\maketitle

\begin{abstract}
The widespread use of vector graphics creates a significant demand for vectorization methods.
While recent learning-based techniques have shown their capability to create vector images of clear topology, filling these primitives with gradients remains a challenge.
In this paper, we propose a segmentation-guided vectorization framework to convert raster images into concise vector graphics with radial gradient fills.
With the guidance of an embedded gradient-aware segmentation subroutine, our approach progressively appends gradient-filled Bézier paths to the output, where primitive parameters are initiated with our newly designed initialization technique and are optimized to minimize our novel loss function.
We build our method on a differentiable renderer with traditional segmentation algorithms to develop it as a model-free tool for raster-to-vector conversion.
It is tested on various inputs to demonstrate its feasibility, independent of datasets, to synthesize vector graphics with improved visual quality and layer-wise topology compared to prior work.
\keywords{Vectorization \and Segmentation \and Differentiable rendering}
\end{abstract}

\section{Introduction}
\label{sec:intro}

Vector graphics offer great flexibility in digital design as they can be easily edited and arbitrarily scaled. Vectorization, the procedure of converting raster images to vector ones, serves as a second way to creating vector graphics other than designing from scratch where extensive artistic skills are required, and also as a bridge between the rapidly developing raster image generation~\cite{rombach2022high} and the relatively less studied vector generation~\cite{reddy2021im2vec,frans2022clipdraw,carlier2020deepsvg,wang2021deepvecfont,ha2018a}.

Vectorization has been explored for decades with various representations proposed to divide images into non-overlapping regions~\cite{xia2009patch,liao2012subdivision,zhu2022tcb,sun2007image,yang2015effective}.
Despite their capability of generating vivid vectorization of realistic images, the complex primitives and lack of hierarchy make the vector output less intuitive for manipulation.

Recent learning-based methods show their potential in preserving image hierarchy, where a vector image is often considered as a list of primitives and their order indicates the structure of the input.
Many deep learning approaches have been proposed to vectorize simple inputs~\cite{reddy2021im2vec} or images of a specific field of interest~\cite{egiazarian2020deep,su2023marvel,reddy2021multi,liu2023dualvector}, but their dependency on models limits them to a particular domain and are not trivially generalizable.
LIVE~\cite{ma2022towards}, in contrast, utilizes DiffVG~\cite{li2020differentiable}, a differentiable renderer, to present a model-free vectorization framework.
It progressively translates a raster image into an SVG in a layer-wise hierarchy, through which the topology of the input is preserved within the order of geometric primitives.
However, its lack of support for gradients results in excessive primitives being added in case of images with rich gradient effects, and adding such support is not as simple as replacing RGBA colors with gradient parameters, as elaborated in \cref{sec:seg_weight}. 

In this paper, we propose a novel segmentation-guided vectorization framework that extends the capability of LIVE to support radial gradients.
The additional parameters of a radial gradient pose an increased challenge to optimization, where an effective method to determine whether a pixel contributes to a path's gradient fill is necessary.
The key insight behind our idea is the similarity between finding contributing pixels and segmentation tasks, based on which we designed a segmentation-guided initialization procedure to progressively append new shapes to the vector output, and a novel loss function to optimize their geometric and gradient parameters.
We evaluated our framework on several datasets with quantitative metrics and a user study, to show its effectiveness and superior performance compared to previous work.

To summarize our contributions:
\begin{itemize}
    \item We introduce a segmentation-guided vectorization framework to create vector graphics automatically with layer-wise hierarchy and radial gradients.
    \item We propose a gradient-aware segmentation method to evaluate the pixel-wise contribution to the geometric and gradient parameters of a path.
    \item We take the segmentation as guidance for our new initialization technique and as a part of our novel segmentation-guided loss.
\end{itemize}

\section{Related Work}
\label{sec:related}

\subsection{Image Vectorization}
Most traditional vectorization methods aim to create vector images of high fidelity with different representations, which could be roughly categorized into mesh-based ones and curve-based ones.
The former representations divide the input image into non-overlapping 2D patches across which colors are interpolated~\cite{tian2022survey}. Shapes of patches include triangular~\cite{xia2009patch,liao2012subdivision,zhu2022tcb}, rectangular~\cite{sun2007image,baksteen2021mesh} or irregular ones such as bézigons~\cite{yang2015effective}. The different selection of mesh shapes determines how patches are organized and how colors are interpolated within patches.
The curved-based representations decompose the image into curves. Diffusion curves, for instance, use curves as geometric primitives with colors defined on sides of the curves~\cite{orzan2008diffusion,xie2014hierarchical}.
While these methods can yield near-photo-realistic results, they may fall short when it comes to ease of editing due to the complex primitives and loss of topological information, compared to a plain list of simple primitives like how an SVG organizes a vector image.
Recent learning-based vectorization work mainly takes the list-of-primitive approach, where the ordered primitives are seen as a sequence of drawing operations, and sequential prediction methods are applied to synthesize vector graphics.
Models are used including recurrent neural networks~\cite{ha2018a,reddy2021im2vec}, transformers~\cite{carlier2020deepsvg}, and are often combined with variational autoencoders~\cite{ha2018a,lopes_learned_2019,kingma2014vae,reddy2021im2vec}.
DiffVG~\cite{li2020differentiable} as a differentiable renderer fills the gap between raster and vector graphics, with which loss functions on raster images could be used directly to optimize the vector images.
A vectorization method optimizing all paths at once is also proposed in the work.

\subsection{Image Topology}
When synthesizing vector graphics, both similarity to the original image and correct topology are important for the ease of subsequent human manipulation.
A related problem is layer decomposition, where images are decomposed into semi-transparent layers. With these layers being vector paths, such a method could serve as a layered vectorization approach. An interactive method~\cite{richardt2014vectorising} is among the first proposed to convert bitmaps into layered vector graphics. Photo2ClipArt~\cite{favreau2017photo2clipart} and other similar work~\cite{du2023image} replace the heavy manual interaction with a user-provided segmentation input, from which the decomposition could be automatically determined.
However, these methods require a concise segmentation for efficient and effective vectorization into a relatively small number of paths, where automatic segmentation algorithms struggle to serve the purpose~\cite{favreau2017photo2clipart}.

Many learning-based methods formulate vector synthesis as sequential prediction problems, through which the order of primitives is naturally preserved.
Some methods predict the primitive parameters directly. For example, Egiazarian et al.~\cite{egiazarian2020deep} use a transformer-based network to directly generate the vectorization of technical line drawings; MARVEL~\cite{su2023marvel} applies deep reinforcement learning to predict strokes in black-and-white comics.
Meanwhile, some approaches use variational autoencoders (VAE)~\cite{kingma2014vae} to create vector graphics.
SketchRNN~\cite{ha2018a} is the first to introduce a Long Short-Term Memory (LSTM)~\cite{hochreiter1997long} based VAE to the representation of vector sketches.
The later DeepSVG~\cite{carlier2020deepsvg} proposed a transformer-based VAE network that also encodes a vector image into a latent code, which could be decoded to reconstruct the vector paths.
SVG-VAE~\cite{lopes_learned_2019} is the first to use raster images as the input to its encoder, thus it could be used to vectorize images.
Im2Vec~\cite{reddy2021im2vec} also takes raster images as its input but drops the need for vector supervision with the help of DiffVG~\cite{li2020differentiable}.
While these methods show great potential in creating topology-aware vector graphics, they rely on specific training datasets and may overlook the fine details in a complex input.
The most similar work to ours is LIVE~\cite{ma2022towards}, in which a progressive framework is proposed to convert raster images into layered vector paths utilizing DiffVG for optimization, but its missing support for color gradients is not trivially achievable by simply optimizing gradient parameters.
Our novel segmentation-guided method takes a step further to handle gradient effects properly, at the same level of simplicity as LIVE, requiring neither additional user input nor deep models.

\section{Method}
\label{sec:method}

\subsection{Method Overview}

\begin{figure}
    \centering
    \includegraphics[width=\linewidth]{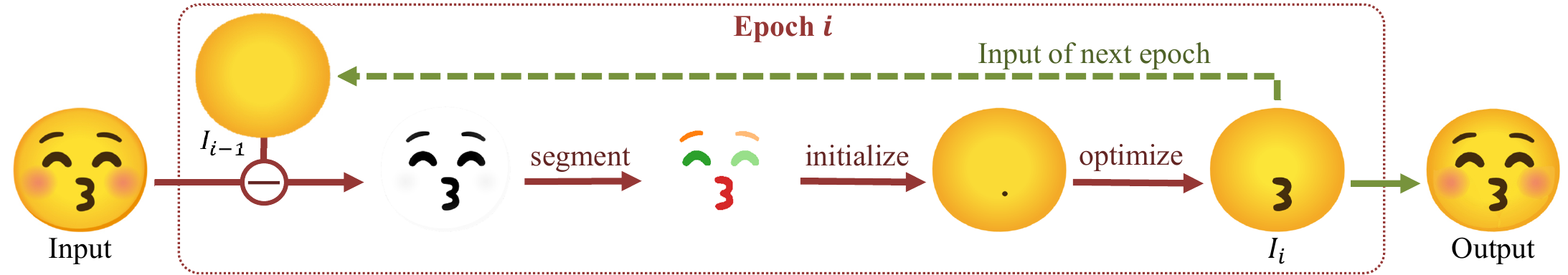}
    \caption{Overview of our framework. \textit{U+1F61A} from Noto Emoji~\cite{noto} is used for demonstration.}
    \label{fig:framework}
\end{figure}

We present a new framework to generate SVG figures from rasterized images, with support for radial gradient fills via a segmentation-guided initialization and optimization process.

Our framework works progressively, where at each epoch, single or multiple Bézier paths are added and optimized. \cref{fig:framework} provides an overview of our method. At the beginning of each epoch $i$, we calculate the difference between the input raster image and the output $I_{i-1}$ from the previous epoch. We segment it with our gradient-aware segmentation (\cref{sec:segmentation}), from which $n_i$ segmented regions are selected for initialization of new paths (\cref{sec:seg_guided_init}). All added paths, including those added in previous epochs, are optimized to minimize the vectorization loss (\cref{sec:seg_weight}). The parameters to be optimized consist of geometric parameters including positions of curve control points and gradient parameters including their center, radius, and color stops. $n_i$, or the number of paths to be appended at $i$-th epoch is specified by the user along with the number of iteration epochs. 

\subsection{Gradient-aware Segmentation}
\label{sec:segmentation}

\begin{figure}
  \centering
  \begin{subfigure}{.15\linewidth}
    \includegraphics[width=\linewidth]{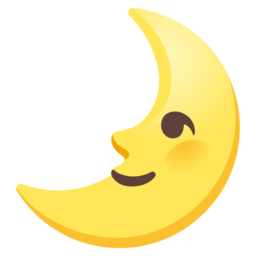}
    \caption{Input}
  \end{subfigure}
  \begin{subfigure}{.15\linewidth}
    \includegraphics[width=\linewidth]{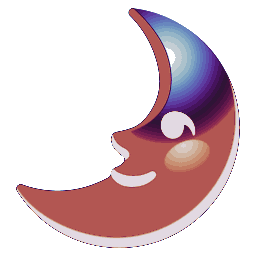}
    \caption{LIVE}
  \end{subfigure}
  \begin{subfigure}{.15\linewidth}
    \includegraphics[width=\linewidth]{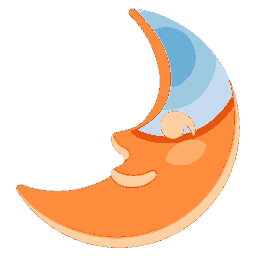}
    \caption{LIVE-30}
  \end{subfigure}
  \begin{subfigure}{.15\linewidth}
    \includegraphics[width=\linewidth]{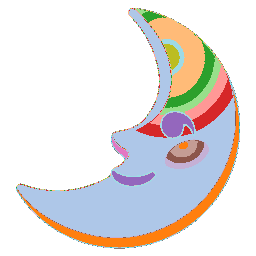}
    \caption{Mean shift}
  \end{subfigure}
  \begin{subfigure}{.15\linewidth}
    \includegraphics[width=\linewidth]{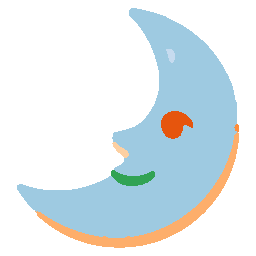}
    \caption{Ours}
  \end{subfigure}
  \caption{Comparisons on segmentation methods. (a) is the raster input. (b) is by LIVE under default settings clustering colors into 200 bins. (c) is also by LIVE but with the number of bins set to 30. (d) is segmented using the Mean-shift~\cite{comaniciu2002mean} algorithm. (e) is by our method.}
  \label{fig:segmentation_method_comparison}
\end{figure}

Quality of path initialization is the key to effective vectorization, hence we present a gradient-aware segmentation method to better estimate the area that a path is likely to span over. Its result is used in the subsequent operations including initialization and optimization (\cref{sec:seg_guided_init}, \cref{sec:seg_weight}).

When paths are filled with a solid color, a connected component of similar or identical colors has a good chance of being covered by one path. LIVE~\cite{ma2022towards} designed a component-wise initialization method, where pixels are clustered into buckets based on their l2-length over RGB channels, and connected pixels in the same bucket are considered one component.
As gradients are considered, the clustering algorithm used by LIVE may result in an excessive segmentation, as shown by \cref{fig:segmentation_method_comparison}.
Other clustering methods including Mean-shift~\cite{comaniciu2002mean} also suffer from over-segmentation for the same reason that colors within a path may vary more than colors between paths.

\begin{figure}[t]
    \centering
    \includegraphics[width=.96\linewidth]{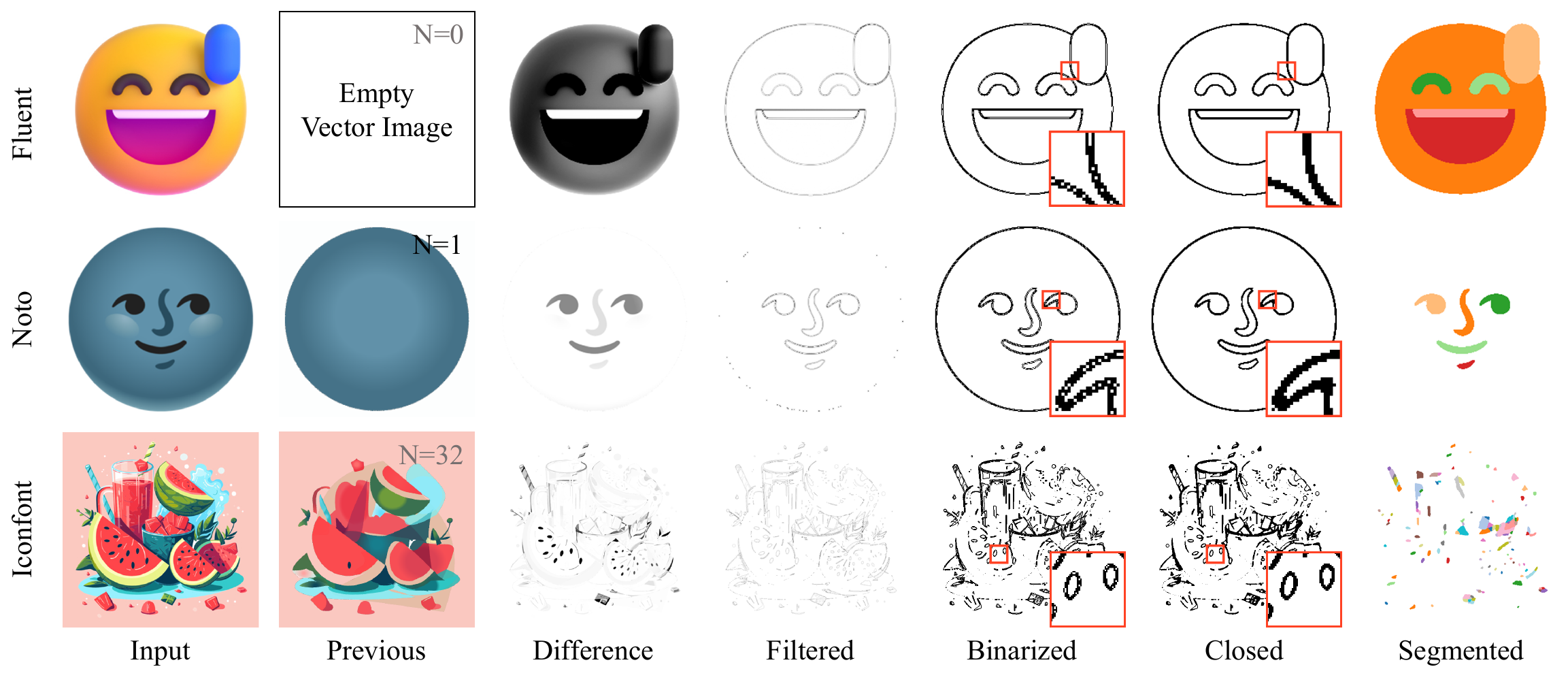}
    \caption{Step-by-step outputs of our gradient-aware segmentation}
    \label{fig:segmentation}
\end{figure}

With the limitations of previous methods, our approach is designed to detect edges of gradients. Colors should derive smoothly inside a region filled with the same gradient fill and are likely to change abruptly at its boundary. Thus, we calculate the secondary spatial gradient with a discrete Laplacian filter
$L=\left[\begin{smallmatrix}1&1&1\\1&-8&1\\1&1&1\end{smallmatrix}\right]$
to identify such rapid changes at gradient boundaries. Our segmentation method features the following steps:

\begin{itemize}
    \item The cross-correlation $S_0\in\mathbb{R}_{w\times h\times 3}$ between the difference to the input and the 2D Laplacian filter is calculated as
    \[S_0=\mathrm{correlate}((I - \hat I)\mathbf{1}_{||\hat I - I||_2 > \epsilon}, L)\,,\]
    where $\hat I$ is the target and $I$ is the synthesized image. Pixels with an error below $\epsilon=0.1$ are excluded.
    \item $S_0$ is then summed over its RGB channels for a grayscale image $S_1\in\mathbb{R}_{w\times h}$, where each pixel of $S_1$ has a value of
    $(S_1)_{ij}=\sum_{c=1}^3\left|(S_0)_{ijc}\right|$.
    \item The grayscale image $S_1$ is converted to a binary image $S_2=\mathbf{1}_{S_1>\mathrm{Otsu}(S_1)}$ with the threshold determined via Otsu's method~\cite{otsu1979threshold}.
    \item Morphological closing~\cite{haralick1987image} and the watershed algorithm~\cite{beucher1993morphological} are then applied to $S_{2}$ for the final segmentation $S\in \mathbb{N}_{h\times w}$.
\end{itemize}

\begin{figure}[t]
    \includegraphics[width=.95\linewidth]{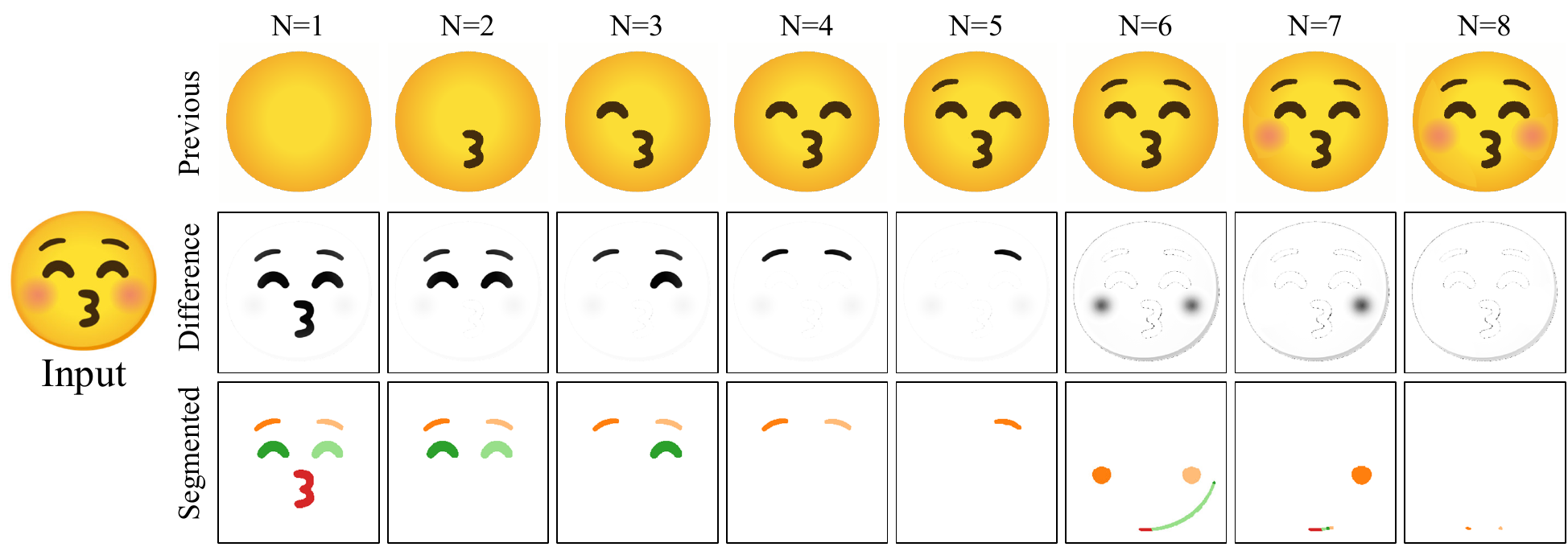}
    \caption{With paths being progressively added, the overall difference between the current output and the raster input decreases, thus we apply Otsu's method~\cite{otsu1979threshold} to automatically determine the threshold for binarization. In this example, the red cheeks get segmented after more significant differences including eyes and brows are fitted, thanks to the dynamic threshold. The differences are normalized and mapped to grayscale.}
    \label{fig:dynamic_threshold}
\end{figure}

In \cref{fig:segmentation} we showcase some intermediate results after each step. Since we are segmenting the difference between the output and the target, the already fitted pixels are ignored and the automatically determined threshold via Otsu's method decreases in response to the descending overall difference, as shown in \cref{fig:dynamic_threshold}. Compared to a fixed threshold, our method makes fewer assumptions about the input and avoids hyperparameters.

\subsection{Segmentation-guided Initialization}
\label{sec:seg_guided_init}

Our vectorization method works in a progressive manner, throughout which we add one or more paths at each epoch. These newly added paths are initialized using our segmentation $S$ as the guidance. For each path to be added, we select the segmentation region with the largest accumulated square error calculated as
\begin{equation}
    w_{i}=\sum_{p\in \Tilde{S}[i]}||I_p-\hat I_p||^2
\end{equation}
where p is iterated over all pixels in $i\text{-th}$ region. This approach prioritizes larger regions to encourage a hierarchical initialization order and prevents regions that are almost properly filled from being chosen. A circle path of four cubic Bézier curves~\cite{ma2022towards} is added at the selected region's center of mass $p_m$. We fill the path with a radial gradient, which centers at $p_m$, with a diameter equal to the geometric mean of the width and the height of the region's bounding box clipped to $[0.2, 1.0]$. The two stop colors at offsets 0\% and 100\% are both initialized to the color of the input image $I$ at ${p_m}$.

\subsection{Loss Function with Segmentation-guided weight}
\label{sec:seg_weight}
\begin{figure}[bt]
    \centering
    \includegraphics[width=.9\linewidth]{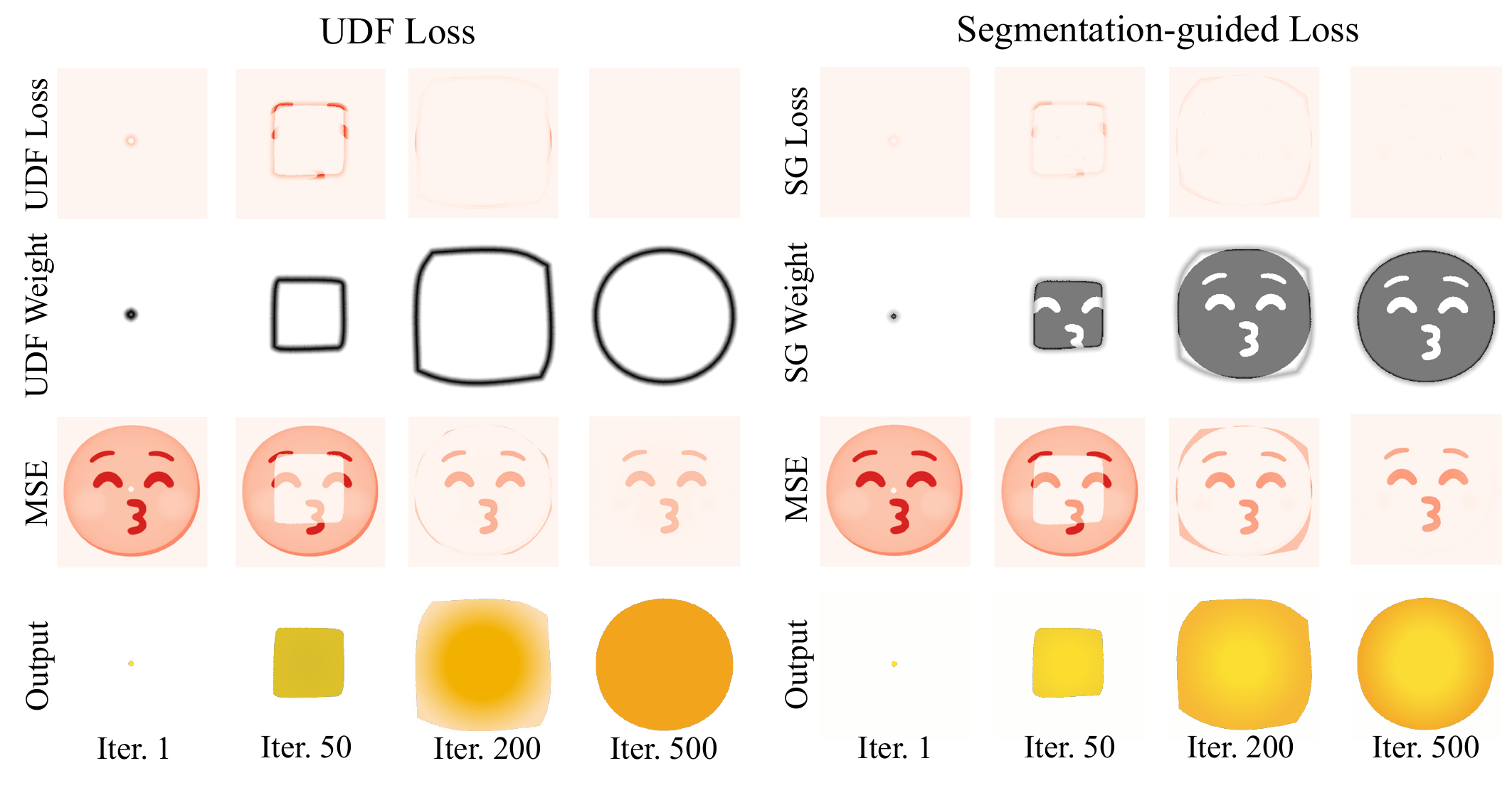}
    \caption{Optimizing course for our method in comparison with UDF loss from LIVE. With our proposed segmentation-guided weight (SG Weight), the gradient fill is optimized to minimize the color error not only on the contour as UDF weight focuses on but also on colors inside the path, while excluding pixels occluded by eyes, brows, and the mouth.}
    \label{fig:seg_weight_comp}
\end{figure}

A common approach to measuring the difference between the synthesized image $\hat I$ and the target image $I$ is through mean squared error (MSE).
However, simply minimizing MSE results in colors biasing towards the average color inside a path, as suggested by LIVE~\cite{ma2022towards}. 
LIVE tackles the problem with its designed UDF loss to focus on pixels on the contour, but for radial gradients, correct colors on the contour do not mean correctness inside, as indicated by \cref{fig:seg_weight_comp}. When gradients are optimized with UDF loss, colors at the center are not fitted well, since those pixels are ignored by the UDF weight.
We draw from the insight by LIVE emphasizing the significance of pixels on edges and extend this concept to include all pixels within a path, except for those occluded by other paths, which is well estimated through our gradient-aware segmentation.

We formulate our segmentation-guided loss with this estimation of unoccluded pixels within a path.
For each added path $p_i$, given the set of pixels covered by the path and the set of pixels within the segment from which the path is initialized, we take their intersection and mark these pixels as being focused. The union of all focused pixels forms a set $F$. We define the weight $w_{\text{SG}}$ as:
\begin{equation}    
    w_\text{SG}(i)=\begin{cases}
    \max(d_i', \alpha_s) ,&i \in F\\
    d_i'(1-\alpha_s) ,&\text{otherwise}
    \end{cases},
    \label{eq:seg_weight}
\end{equation}
where $i$ is an index of pixel and $d_i'$ is the UDF weight from~\cite{ma2022towards}. The UDF weight is introduced to give the pixels near the path contour a higher weight, as depicted in \cref{fig:seg_weight_comp}.
$\alpha_s$ balances between the UDF weight $d_i'$ and our segmentation weight, and is set to $0.6$ empirically.
$d_i'$ is defined with:
\begin{equation}
    {d}'_i= \frac{\mathrm{ReLU}(\tau -\left | d_i \right |)}{\sum_{j=1}^{w\times h} \mathrm{ReLU}(\tau -\left | d_j \right |)},
\label{eq:sdf_loss_d}
\end{equation}
where $d_i$ is the distance from the pixel $i$ to the nearest path contour, which is then normalized and saturated with threshold $\tau=10$.

We re-weight the pixel-wise color error with our proposed weight $w_{\text{SG}}$ for the final segmentation-guided loss:
\begin{equation}
    \mathcal{L}_{\text{SG}}=\frac13\sum_{i=1}^{w\times h}w_{\text{SG}}(i)\sum_{c=1}^3(I_{c,i}-\hat I_{c,i})^2,
\end{equation}
where $i$ is iterated over all pixels and squared color differences are averaged over the RGB channels.

\begin{figure}[t]
    \centering
    \includegraphics[width=1.0\linewidth]{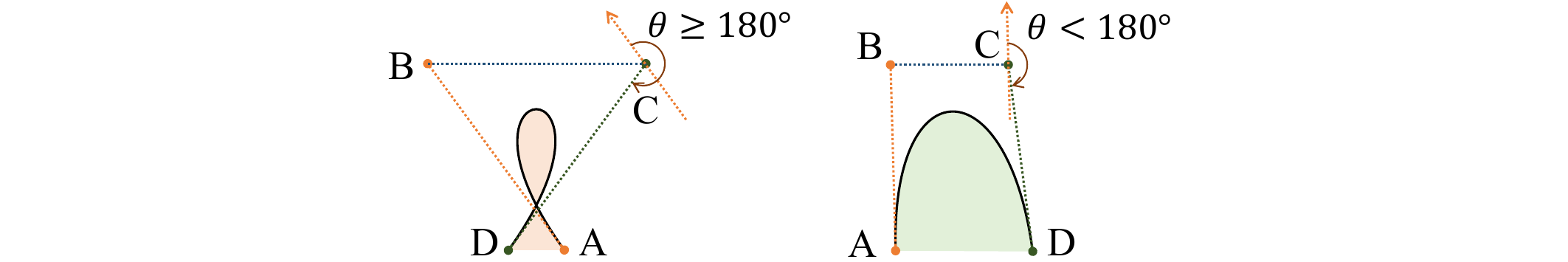}
    \caption{Bézier paths with four control points}
    \label{fig:xing}
\end{figure}

We also introduce Xing loss~\cite{ma2022towards} to help relieve self-intersection. Given a cubic Bézier curve with four control points $A,\,B,\,C,\,$ and $D$, self-intersection is more likely to occur when the angle between $\vec{AB}$ and $\vec{CD}$ is greater than 180°, as shown in \cref{fig:xing}. The Xing loss $\mathcal{L}_{\mathrm {Xing}}$ is defined to penalize angles over 180°:
\begin{equation}
    D_1=\begin{cases}
    1,\,\vec{AB}\times\vec{BC}>0\\0,\,otherwise\end{cases},\,D_2=\frac{\vec{AB}\times\vec{CD}}{\left\Vert\vec{AB}\right\Vert\,\left\Vert\vec{CD}\right\Vert}
\label{eq:xing_loss}
\end{equation}
\begin{equation}
    \mathcal{L}_{\text{Xing}}=D_1(\text{ReLU}(-D_2))+(1-D_1)(\text{ReLU}(D_2)).
\end{equation}

Our optimization objective is defined as follows:
\begin{equation}
\mathcal{L}=\mathcal{L}_{\text{SG}}+\lambda\mathcal{L}_{\text{Xing}}, 
\end{equation}
where $\lambda$ is set to 0.05 empirically.

\section{Experiments}
\label{sec:Experiments}

\subsection{Implementation Details}
\label{sec:Implementation}

We implement our framework with DiffVG~\cite{li2020differentiable} renderer based on PyTorch. Adam optimizer\cite{kingma2014adam} is used with a learning rate of $10^{-2}$ and 1 for gradient parameters and path points respectively. We use scikit-image~\cite{van2014scikit} for its implementation of morphological operations and the watershed algorithm~\cite{beucher1993morphological}.

\begin{figure}
    \centering
    \includegraphics[width=.8\linewidth]{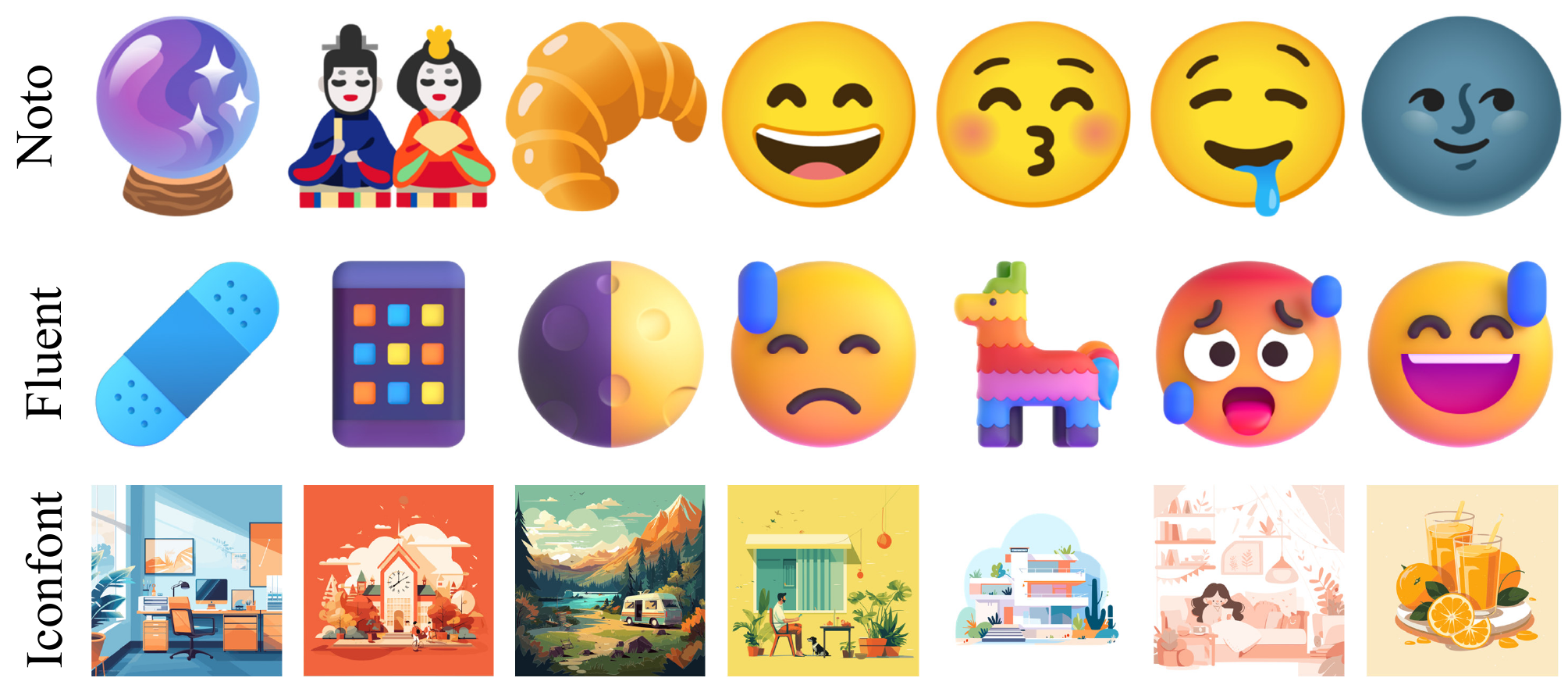}
    \caption{Exemplars from our chosen datasets for evaluation}
    \label{fig:datsets}
\end{figure}

\subsection{Datasets}
\label{sec:datasets}

\paragraph{Noto Emoji} We randomly select 256 emojis from Noto Emoji~\cite{noto}(Unicode 15.0) varying in colors and genres. In contrast to previous work~\cite{ma2022towards, reddy2021im2vec}, we take a more recent version of the dataset, consisting of resigned emojis filled with gradients instead of single colors. Emojis from this dataset consist of a relatively small number of paths with a clear hierarchy, thus we evaluate on this dataset to show our method's capability to decompose images into layers with gradients.

\paragraph{Fluent Emoji} We select an additional 256 images at random from Microsoft's Fluent Emoji~\cite{fluent} collection. This dataset features colorful emojis with rich gradient details. We test on this dataset to elaborate on our method's ability to match up with target images using fewer paths compared to previous work~\cite{ma2022towards, li2020differentiable}.

\paragraph{Iconfont} A set of 128 vector arts are fetched from the online Iconfont Illustration Library~\cite{iconfont}. Images in the dataset contain an average of 1020 paths with great details but no gradients. We use these images to exhibit that our segmentation-guided approach is also suitable for vectorizing complex inputs, and performs on par with or better than previous work even without gradients.

It is worth noticing that our framework is model-free, thus datasets mentioned above are for sole evaluation purposes.

\begin{figure}
    \centering
    \includegraphics[width=.48\linewidth]{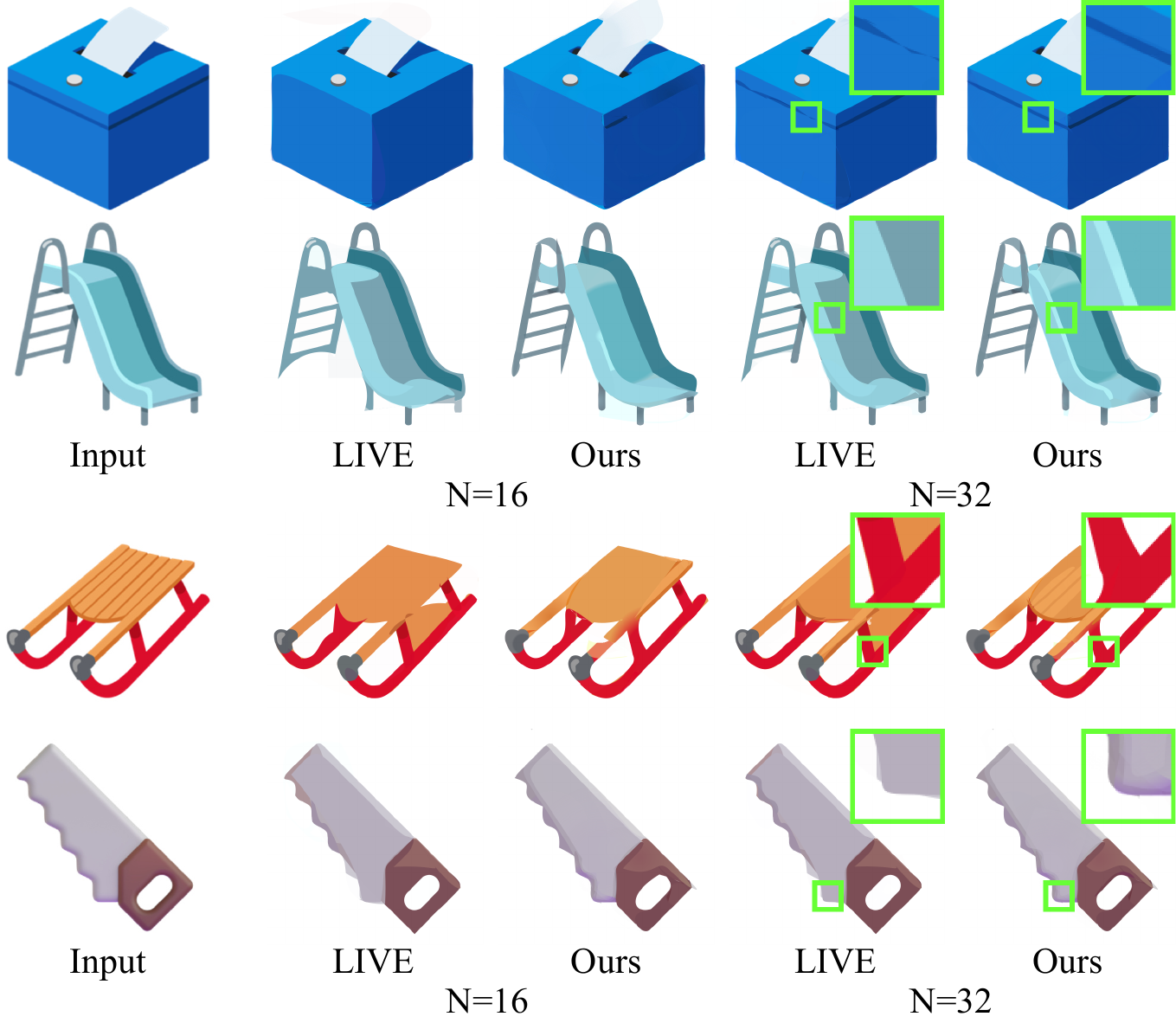}
    \hspace{.02\linewidth}
    \includegraphics[width=.48\linewidth]{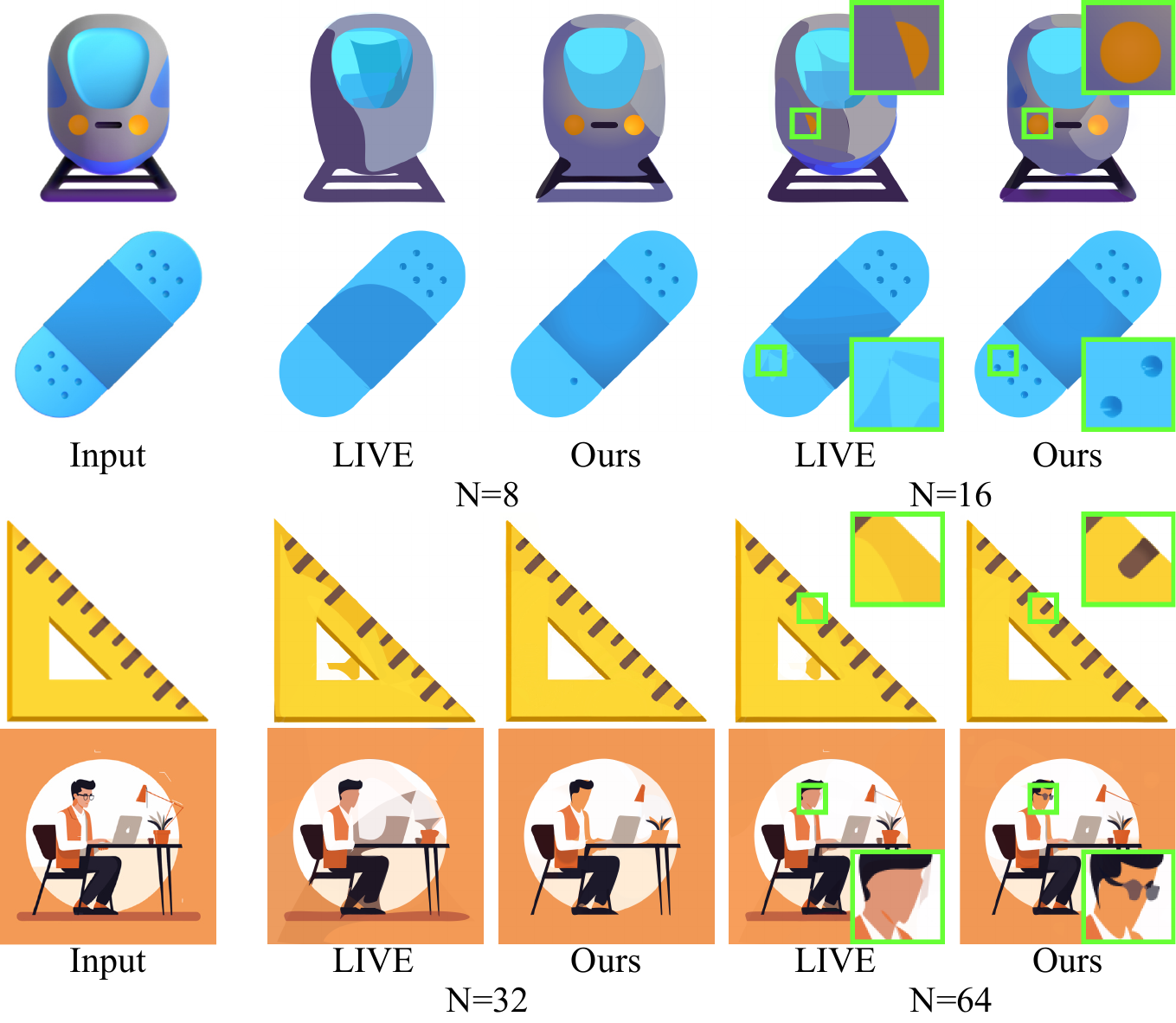}
    \caption{Qualitative comparisons with LIVE}
    \label{fig:more}
\end{figure}

\subsection{Qualitative Comparison}
With our initial goal of supporting gradient fills in a topology-preserving vectorization approach, we conducted a comparative analysis of the visual quality of our reconstructed vector graphics against LIVE~\cite{ma2022towards}. Our method achieves superior visual quality in vectorization while utilizing the same number of paths as LIVE, as evidenced in \cref{fig:more}.

\begin{figure}
    \centering
    \includegraphics[width=\linewidth]{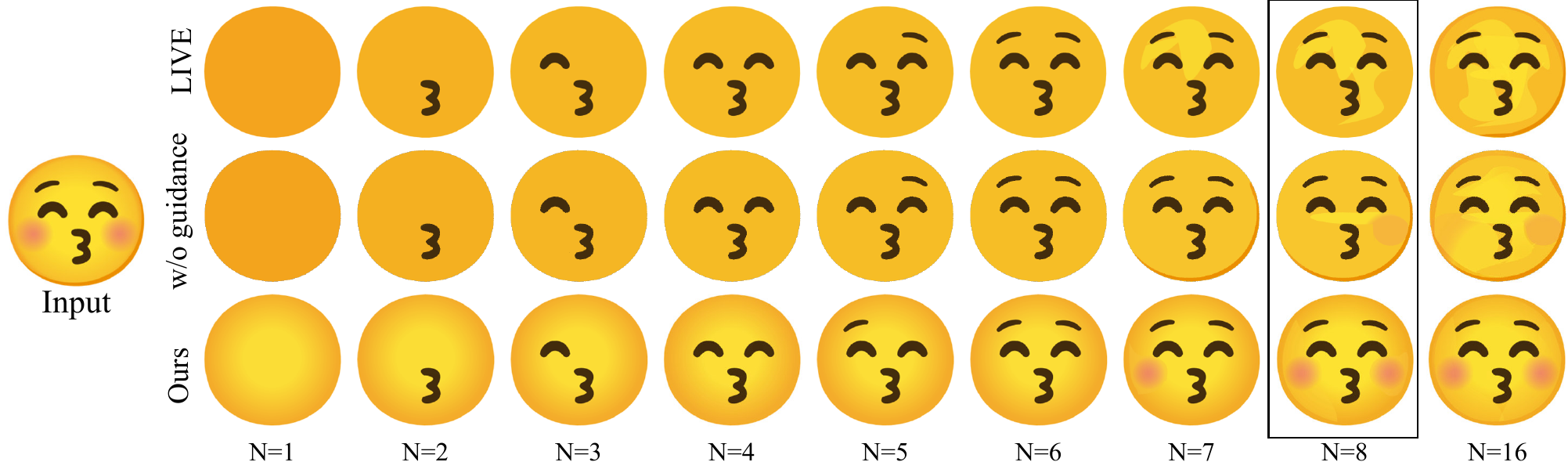}
    \includegraphics[width=\linewidth]{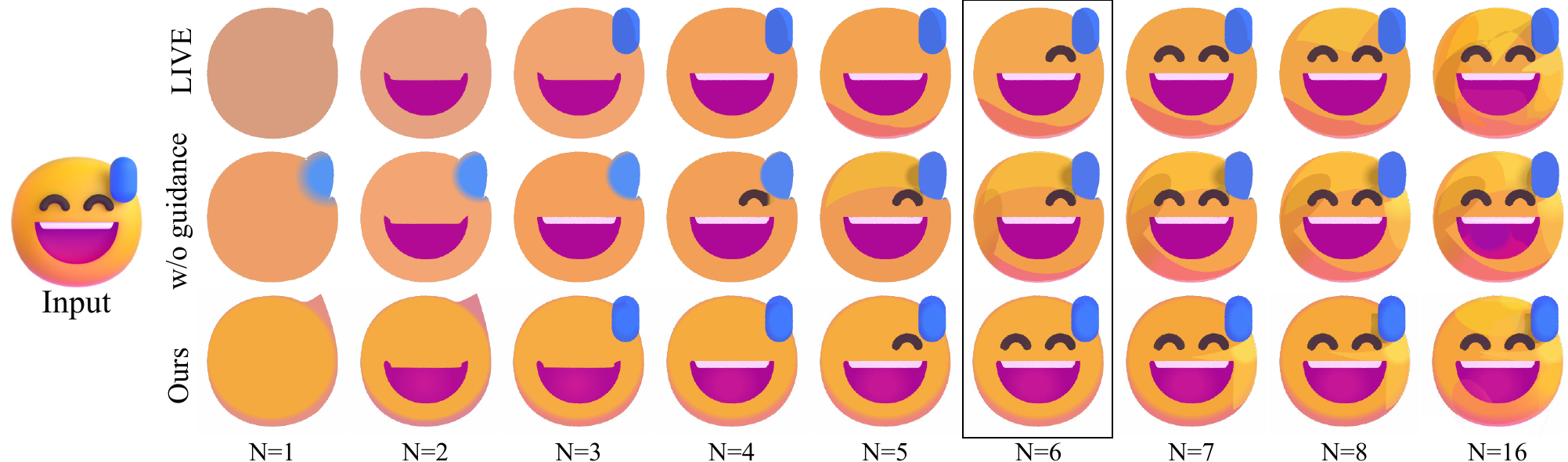}
    \caption{A comparison on number of paths needed to reconstruct major components. `w/o guidance' refers to gradient fills being added without our proposed segmentation guidance. For the first input (\textit{U+1F61A} from Noto Emoji), our method vectorizes all elements with 8 paths, while adding more paths does not deteriorate the output. For the second input (\textit{U+1F605} from Fluent Emoji), ours reconstructs the facial parts with 6 paths and achieves a close gradient effect to the input using 16 paths.}
    \label{fig:quality_comparison}
\end{figure}

As depicted in \cref{fig:quality_comparison}, when presented with input containing gradients, LIVE struggles to accurately vectorize gradient-filled facial features, despite correctly capturing other facial parts. In contrast, our approach accurately reconstructs the input using concise paths. Additionally, attempts to optimize gradient parameters without guidance, as seen in the `w/o guidance' rows, result in degraded outcomes.

\begin{figure}
    \centering
    \includegraphics[width=\linewidth]{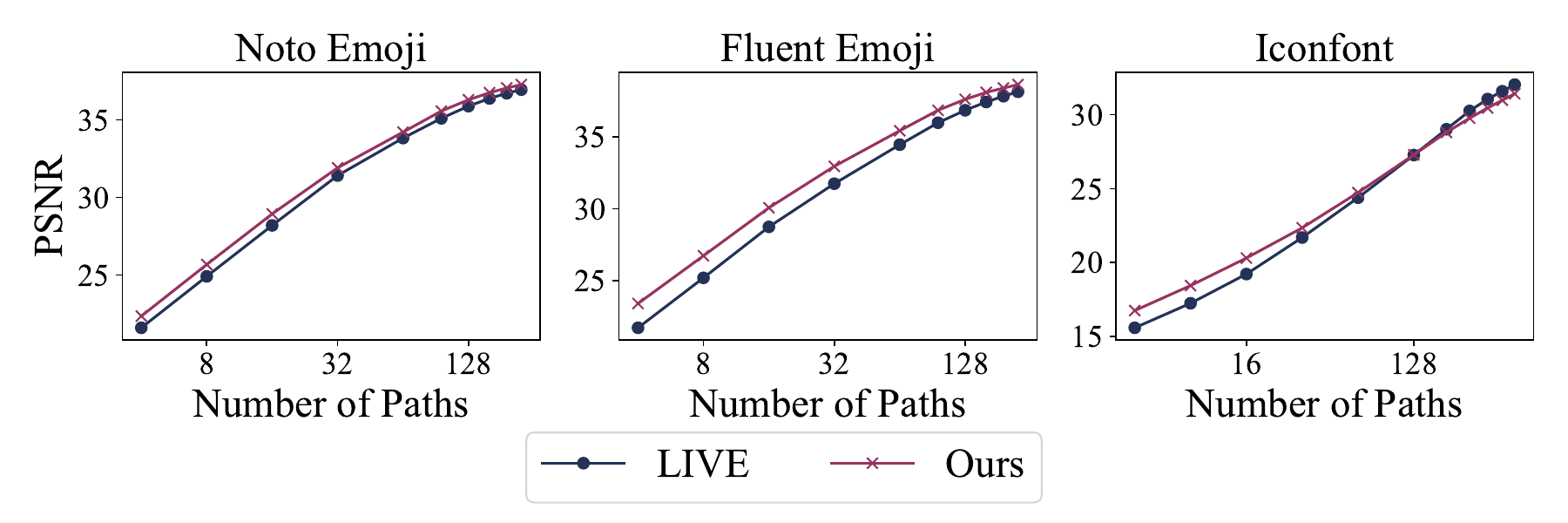}
    \caption{Quantitative comparison between our method and LIVE. Our method achieves significantly lower error when the numbers of paths are small and converge to equal performance with excessive paths being added.}
    \label{fig:mse_psnr_comparison}
\end{figure}

\subsection{Quantitative Comparison}
We further adopt quantitative metrics in a quantitative comparison with LIVE.
We calculated PSNR to measure the difference between the rendered vectorized outputs and their corresponding raster inputs. For Noto and Fluent Emoji datasets, we add $\text{clamp}(2^{i-2}, 1, 32)$ paths at $i$-th initialization to reach a total of 256 paths. For the Iconfont dataset, 512 paths are used for each image with $\text{clamp}(2^{i-2}, 1, 64)$ paths being added at $i$-th epoch, for its higher complexity.

As reported in \cref{fig:mse_psnr_comparison}, our method achieves generally faster convergence than LIVE, especially when a small number of paths are added.
Our segmentation-guided framework can capture large-scale gradient features where LIVE tends to emit superfluous paths.
With excessive paths being added, both methods yield the same level of quality in terms of PSNR. 

\subsection{Layer Decomposition}
\begin{figure}
  \centering
  \begin{subfigure}{.15\linewidth}
    \includegraphics[width=\linewidth]{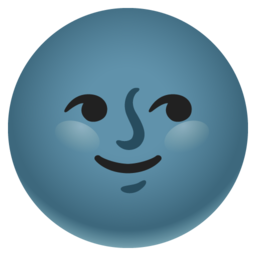}
    \caption{Source}
  \end{subfigure}
  \begin{subfigure}{.15\linewidth}
    \includegraphics[width=\linewidth]{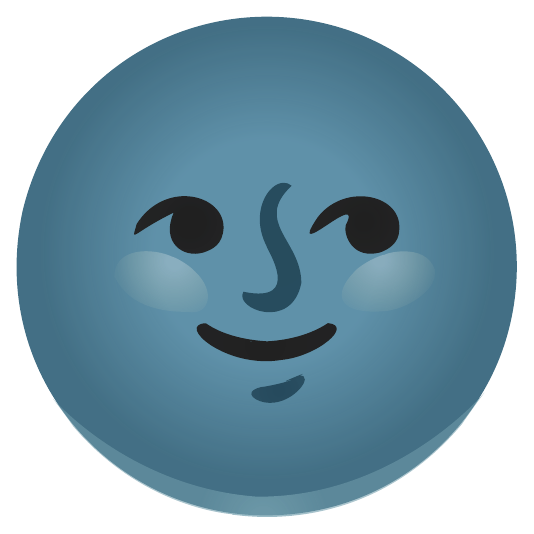}
    \caption{Vectorized}
  \end{subfigure}
  \begin{subfigure}{.30\linewidth}
    \includegraphics[width=\linewidth]{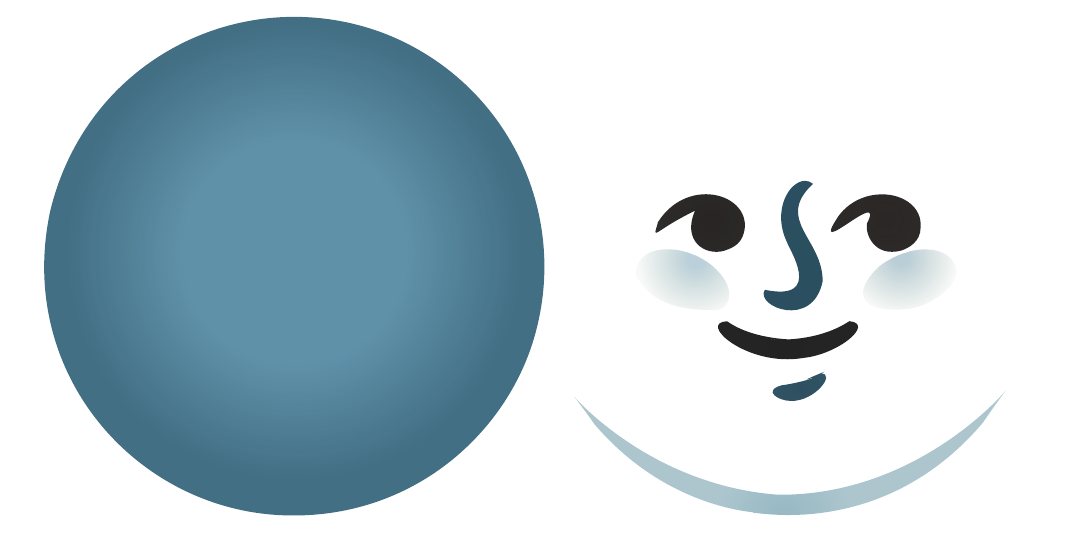}
    \caption{Decomposed layers}
  \end{subfigure}
  \begin{subfigure}{.15\linewidth}
    \includegraphics[width=\linewidth]{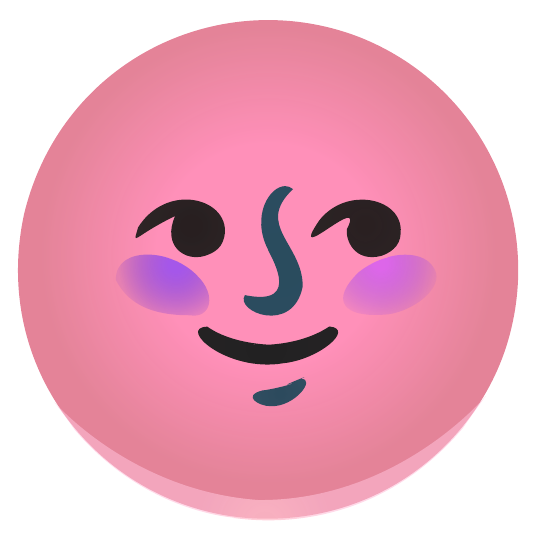}
    \caption{Edited}
  \end{subfigure}
  \caption{Decomposed layers and recoloring}
  \label{fig:layer_decomposition}
\end{figure}

The framework we propose captures layer-wise structure during our progressive vectorization. The segmentation-guided initialization prioritizes segments that are more significant in terms of accumulated error, while details with relatively small errors are captured later by our adaptive gradient-aware segmentation, as shown in \cref{fig:dynamic_threshold}. Given an input with a clear hierarchy, these progressively added ordered paths resemble the handcrafted vector graphics, creating an easy-to-edit output, as shown in \cref{fig:layer_decomposition}.

\subsection{User Study}

\begin{table}[t]
\begin{minipage}[t]{0.48\linewidth}
\caption{Results collected from our user study. The row `Total' stands for the total number of answers collected for all the images in the dataset vectorized given the number of paths. As shown by the results, users in general prefer the vectorization by our method. }
\label{tbl:user_study}

\begin{subtable}{\linewidth}\centering
\caption{Noto Emoji}
\resizebox{\linewidth}{!}{%
\begin{tabularx}{1.1\linewidth}{lXXXXX}
\toprule
\#Paths & 8 & 16 & 32 & 64 & Overall \\
\midrule
LIVE & 39.0\% & 39.0\% & 40.8\% & 43.0\% & 40.4\% \\
Ours & \textbf{61.0\%} & \textbf{61.0\%} & \textbf{59.2\%} & \textbf{57.0\%} & \textbf{59.6\%} \\
Total & 449 & 441 & 441 & 454 & 1785\\
\bottomrule
\end{tabularx}}
\end{subtable}

\vspace*{.1in}

\begin{subtable}{\linewidth}\centering
\caption{Fluent Emoji}
\resizebox{\linewidth}{!}{%
\begin{tabularx}{1.1\linewidth}{lXXXXX}
\toprule
\#Paths & 8 & 16 & 32 & 64 & Overall \\
\midrule
LIVE & 40.2\% & 38.4\% & 30.8\% & 29.3\% & 34.7\% \\
Ours & \textbf{59.8\%} & \textbf{61.6\%} & \textbf{69.2\%} & \textbf{70.7\%} & \textbf{65.3\%} \\
Total & 433 & 445 & 452 & 426 & 1756 \\
\bottomrule
\end{tabularx}}
\end{subtable}

\vspace*{.1in}

\begin{subtable}{\linewidth}\centering
\caption{Iconfont}
\resizebox{\linewidth}{!}{%
\begin{tabularx}{1.1\linewidth}{lXXXXX}
\toprule
\#Paths & 32 & 64 & 128 & 256 & Overall \\
\midrule
LIVE & 32.2\% & 42.7\% & 48.5\% & 44.3\% & 42.1\% \\
Ours & \textbf{67.8\%} & \textbf{57.3\%} & \textbf{51.5\%} & \textbf{55.7\%} & \textbf{57.9\%} \\
Total & 214 & 241 & 227 & 237 & 919 \\
\bottomrule
\end{tabularx}}
\end{subtable}

\end{minipage}\hspace{.04\linewidth}\begin{minipage}[t]{.48\linewidth}
    \caption{Results of our ablation study in PSNR. `G' in the header is for `with gradients' and `S' is for `with segmentation guidance'. The first row, with neither gradients nor segmentation, stands for LIVE.}
    \centering
    \begin{subtable}{\linewidth}\centering
    \caption{Noto Emoji}
    \resizebox{\linewidth}{!}{%
    \begin{tabularx}{1.1\linewidth}{cc|XXXXXX}
        \toprule
        G & S & N=8 & 16 & 32 & 64 & 128 & 256 \\
        \midrule
 &  & 24.92 & 28.21 & 31.42 & 33.83 & 35.89 & 37.16 \\
 & \checkmark & {\ul 25.33} & {\ul 28.67} & 31.63 & 34.13 & {\ul 36.12} & {\ul 37.34} \\
\checkmark &  & 25.19 & 28.51 & {\ul 31.80} & {\ul 34.17} & 36.05 & 37.29 \\
\checkmark & \checkmark & \textbf{25.69} & \textbf{28.95} & \textbf{31.92} & \textbf{34.22} & \textbf{36.30} & \textbf{37.46}
        \\\bottomrule
    \end{tabularx}}
    \end{subtable}

\vspace*{.1in}

    \begin{subtable}{\linewidth}\centering
    \caption{Fluent Emoji}
    \resizebox{\linewidth}{!}{%
    \begin{tabularx}{1.1\linewidth}{cc|XXXXXX}
        \toprule
        G & S & N=8 & 16 & 32 & 64 & 128 & 256 \\
        \midrule
&  & 25.19 & 28.74 & 31.74 & 34.46 & 36.85 & 38.38 \\
& \checkmark & {\ul 26.26} & 29.36 & 32.22 & 34.67 & 36.96 & 38.52 \\
\checkmark &  & 26.15 & {\ul 29.59} & {\ul 32.64} & {\ul 35.28} & {\ul 37.54} & \textbf{38.91} \\
\checkmark & \checkmark & \textbf{26.73} & \textbf{30.07} & \textbf{32.95} & \textbf{35.43} & \textbf{37.61} & {\ul 38.84}
        \\\bottomrule
    \end{tabularx}}
    \end{subtable}

\vspace*{.1in}

    \begin{subtable}{\linewidth}\centering
    \caption{Iconfont}
    \resizebox{\linewidth}{!}{%
    \begin{tabularx}{1.1\linewidth}{cc|XXXXXX}
        \toprule
        G & S & N=16 & 32 & 64 & 128 & 256 & 512 \\
        \midrule
        &  & 19.22 & 21.69 & 24.38 & 27.26 & {\ul 30.25} & {\ul 32.36} \\
        & \checkmark & {\ul 20.02} & {\ul 22.02} & 24.29 & 26.86 & 29.43 & 31.44 \\
       \checkmark &  & 19.42 & 21.96 & {\ul 24.72} & \textbf{27.60} & \textbf{30.67} & \textbf{32.72} \\
       \checkmark & \checkmark & \textbf{20.31} & \textbf{22.35} & \textbf{24.73} & {\ul 27.28} & 29.76 & 31.74
        \\\bottomrule
    \end{tabularx}}
    \end{subtable}

    \label{tbl:ablation_psnr}
\end{minipage}
\end{table}

\begin{figure}
    \centering
    \includegraphics[width=\linewidth]{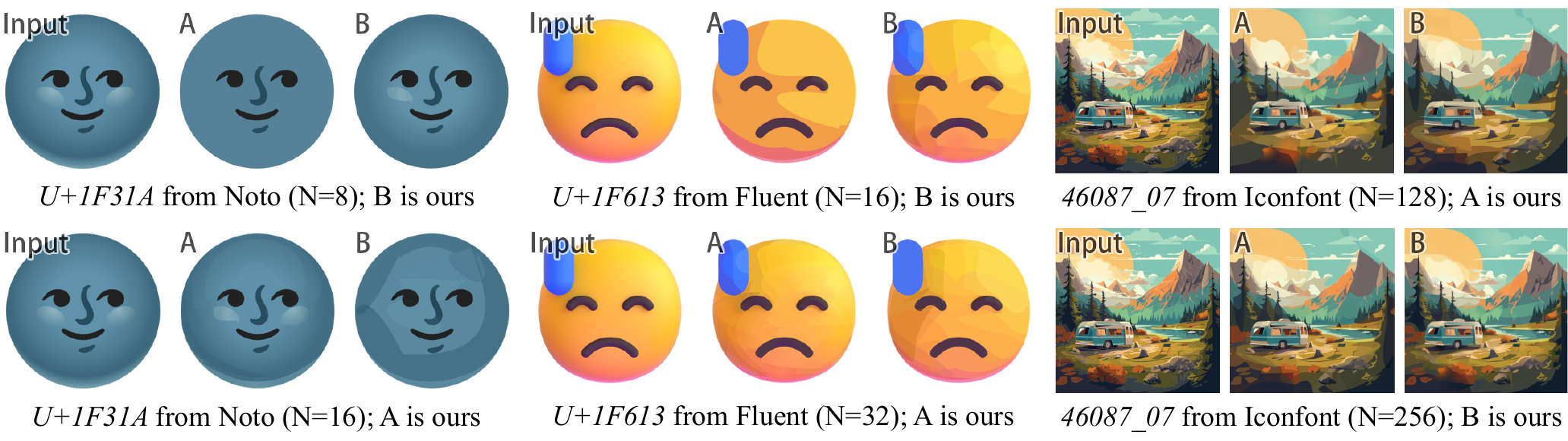}
    \caption{Examples from our questionnaire. Participants are asked to
choose the better one for each pair of outputs.}
    \label{fig:question_sample}
\end{figure}

We conducted a questionnaire survey to obtain subjective feedback from users.
Each participant is presented with 20 questions randomly selected from a total of 2,560 questions, that is, 4 numbers of paths multiplied by 640 images from the three datasets.
In each question, participants are shown a raster input and two vectorized versions of it, one using LIVE and the other using our method, with the same number of paths. They are asked to choose the one that looks closer to the original image or, at an equivalent level of similarity, looks more appealing. \Cref{fig:question_sample} shows examples.

In our study, 223 people engaged, contributing a total of 4,460 votes. The results, presented in \cref{tbl:user_study}, reveal a clear preference for our approach. In the Fluent Emoji dataset, our method is notably preferred for these images with rich gradients. Despite the absence of gradients in many images from Noto Emoji and Iconfont, our method remains the preferred choice.

\subsection{Ablation Study}

\begin{figure}[b]
    \centering
    \includegraphics[width=.49\linewidth]{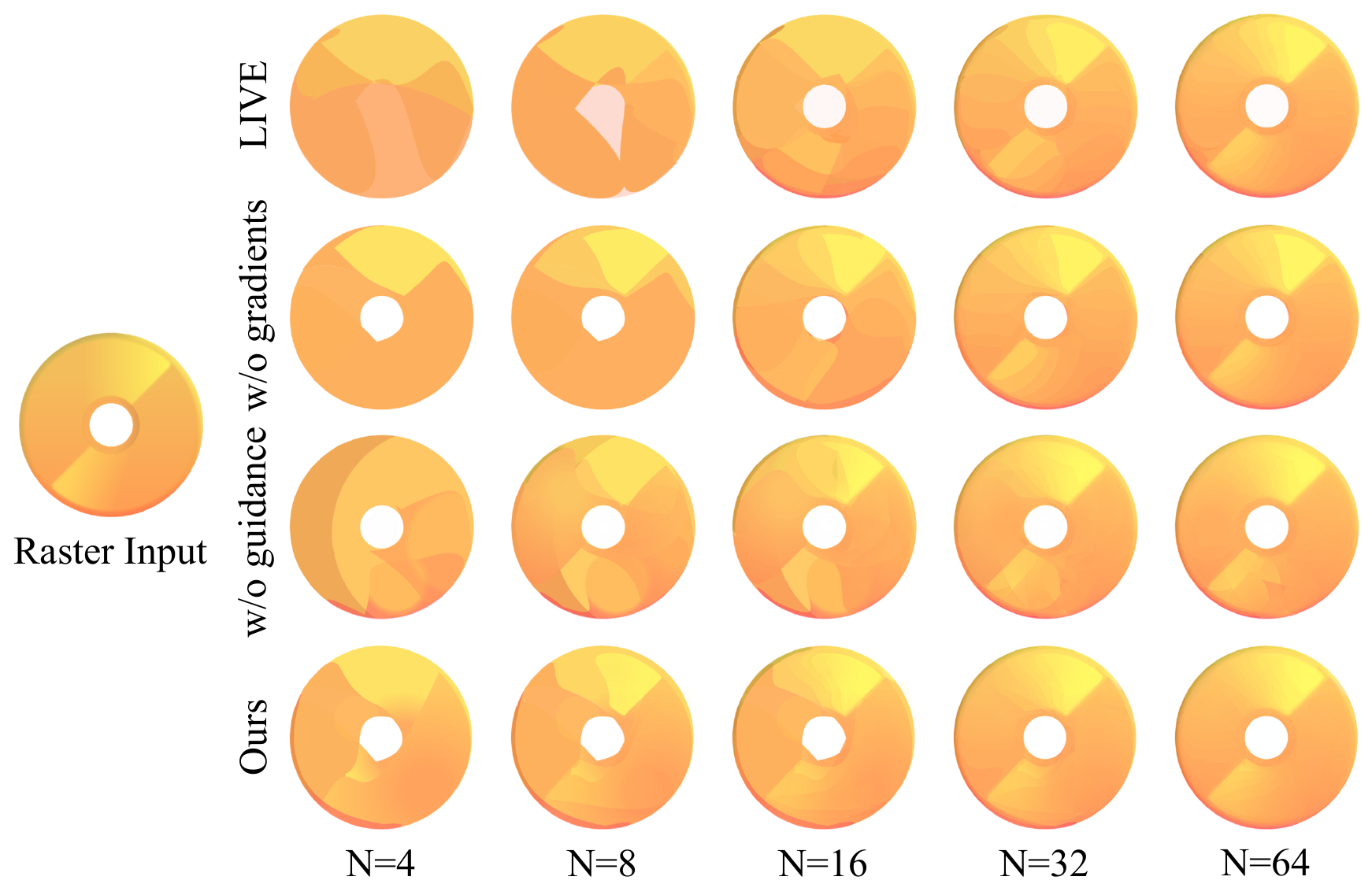}
    \includegraphics[width=.49\linewidth]{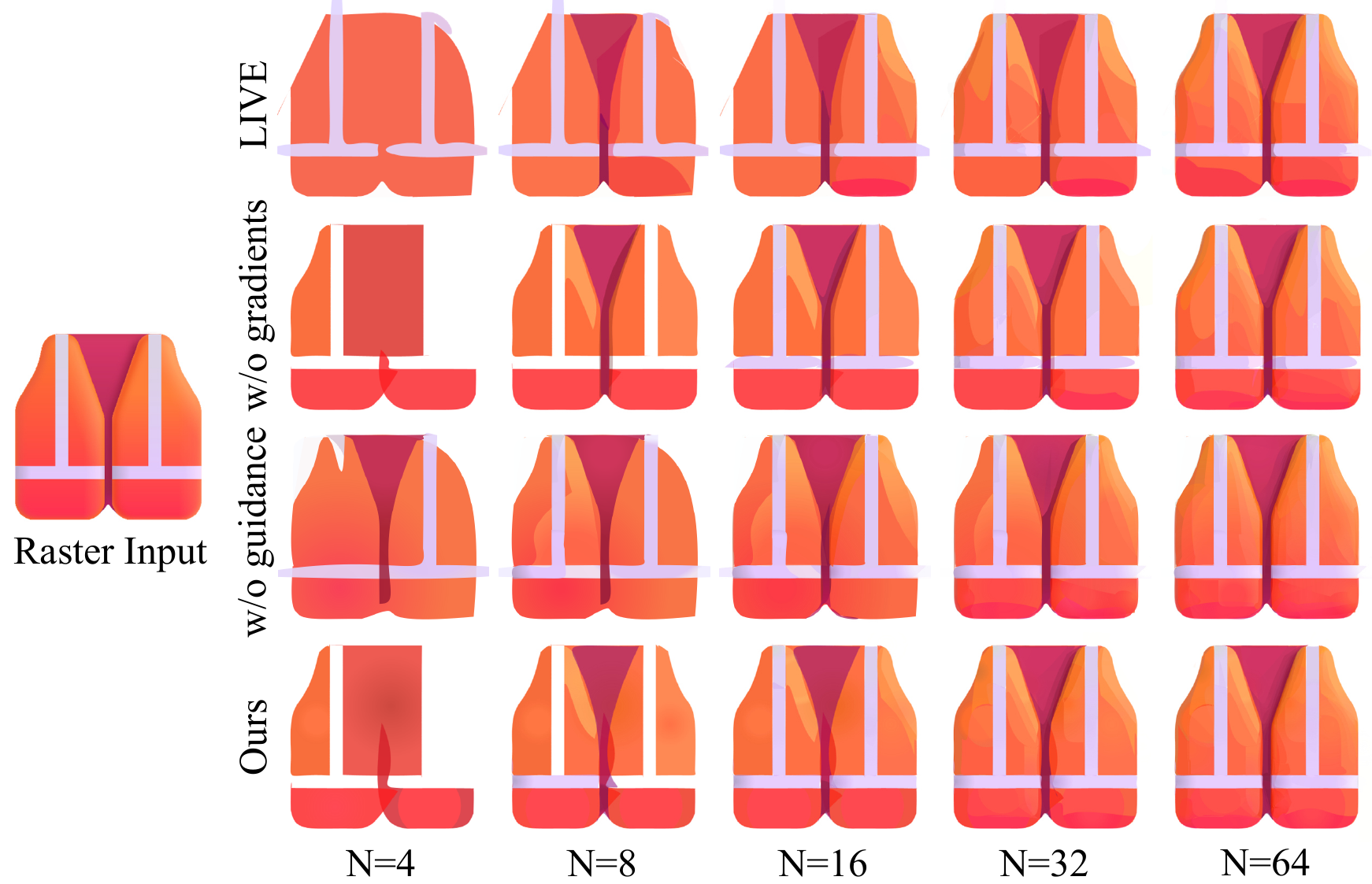}
    \caption{Comparison between four combinations tested in the ablation study. When neither gradients nor segmentation guidance is applied, our method works the same as LIVE (1\textsuperscript{st} row).}
    \label{fig:ablation_study}
\end{figure}

To verify the effectiveness of our contributions, we conducted an ablation study, by vectorizing without gradients, without segmentation guidance, or both, as shown in \cref{fig:ablation_study}. The similarity between vectorized results and their corresponding inputs is measured by PSNR. Results are tabulated in \cref{tbl:ablation_psnr}.

\paragraph{Gradients} For images with rich gradient effects, such as those in Fluent Emoji, the introduction of gradients significantly improves the outcome. For Noto Emoji, where most images do not have gradients, the improvement is less noticeable. Iconfont is an interesting case; although its paths are filled with solid colors, a form of approximate color transition is achieved through multiple paths with progressively changing colors. Introducing gradients also contributes to a noticeable enhancement.

\paragraph{Segmentation guidance} Compared to solely introducing gradients, incorporating segmentation guidance has proven more effective. As illustrated in \cref{tbl:ablation_psnr}, a combination of gradients and segmentation yields superior results on both Noto and Fluent datasets. For the Iconfont dataset, the 
imitated color transition is considered a gradient by the segmentation; thus, with fewer paths, the guided variants achieve a more approximate visual effect through large gradient paths. When more paths are added, using smaller, precise solid color paths yields better results, while the guided vectorization remains competitive in terms of accuracy, and is preferred in the user study.

\subsection{Limitations and Future Work}

\begin{figure}
    \centering
    \includegraphics[width=.75\linewidth]{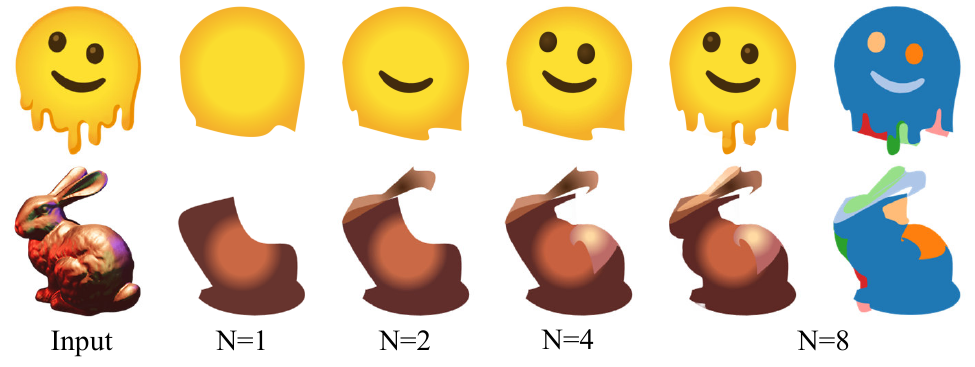}
    \caption{Intricate contours that our simple initial paths fail to converge to. Extra paths are added to mask the overflowing colors.}
    \label{fig:failure_contour}
\end{figure}

While our segmentation guidance effectively captures large-scale features, the initial shape, consisting of a loop formed by four connected Bézier curves, encounters challenges in accurately fitting intricate shapes, as illustrated in \Cref{fig:failure_contour}. Achieving a balance between the number of paths and the number of control points per path needs further deliberation.

Additionally, our implementation employs radial gradients with two color stops for simplicity, since a linear gradient can be interpreted as a radial gradient with its center positioned outside the path. Our method is not confined to a specific type of gradient. A more comprehensive implementation could involve dynamically determining gradient types and adjusting the number of color stops accordingly.

\section{Conclusion}
\label{sec:conclusion}

In this paper, we propose a segmentation-guided layer-wise vectorization framework, which synthesizes vector images by progressively adding paths and filling the paths with radial gradients. We design a gradient-aware segmentation method out of traditional algorithms to guide our novel initialization approach and newly designed loss function to address the obstacles in optimizing gradient parameters. We test our method on Noto~\cite{noto} and Fluent Emoji~\cite{fluent} to show its capability in decomposing raster images into concise layers of gradient-filled paths when the topology is clear, and on Iconfont~\cite{iconfont} to demonstrate its effectiveness in case of complex inputs. Our method shows superior performance compared to previous work.

\section*{Acknowledgements}
This work was supported by the NSFC under Grant 62072271.

%
%
\bibliographystyle{splncs04}
\bibliography{main}
\end{document}